\documentclass[letterpaper, 10 pt, journal, twoside]{IEEEtran}
\usepackage{amsmath,amsfonts}
\usepackage{algorithmic}
\usepackage{algorithm}
\usepackage{array}
\usepackage[caption=false,font=normalsize,labelfont=sf,textfont=sf]{subfig}
\usepackage{textcomp}
\usepackage{stfloats}
\usepackage{url}
\usepackage{verbatim}
\usepackage{graphicx}
\usepackage{cite}
\usepackage{multirow}
\usepackage{color}
\usepackage{marvosym}
\usepackage{ifsym}
\usepackage[T1]{fontenc}
\usepackage{hyperref}
\usepackage{aecompl}
\begin{document}
\title{
Learning Semantic-Agnostic and Spatial-Aware Representation for Generalizable Visual-Audio Navigation}

\author{Hongcheng Wang$^{*1}$, Yuxuan Wang$^{*2}$, Fangwei Zhong$^{3}$, Mingdong Wu$^1$,\\ Jianwei Zhang$^4$, Yizhou Wang$^5$ and Hao Dong$^{\ddag 1}$
\thanks{Manuscript received: December 28, 2022; Revised: March 22, 2023; Accepted: April 18, 2023.}
\thanks{This paper was recommended for publication by Editor Eric Marchand upon evaluation of the Associate Editor and Reviewers’ comments.}
\thanks{*These authors contributed equally.}
\thanks{$\ddag$Corresponding author.}
\thanks{$^{1}$Hongcheng Wang, Mingdong Wu and Hao Dong are with School of Computer Science, Peking University, Beijing, 100871, China. 
        {\tt\small \{whc.1999, wmingd, hao.dong\}@pku.edu.cn}}%
\thanks{$^{2}$Yuxuan Wang is with Academy for Advanced Interdisciplinary Studies, Peking University, Beijing, 100871, China.
        {\tt\small yuxwang@pku.edu.cn}}%
\thanks{$^{3}$Fangwei Zhong is with School of Intelligence and Technology, Peking University and National Key Laboratory of General Artificial Intelligence, BIGAI, Beijing, 100871, China.
        {\tt\small zfw@pku.edu.cn}}
        \thanks{$^{4}$Jianwei Zhang is with TAMS, Department of Informatics, Universität Hamburg, Germany.
        {\tt\small jianwei.zhang@uni-hamburg.de}}
        \thanks{$^{5}$Yizhou Wang is with the Center on Frontiers of Computing Studies, Institute for Artificial Intelligence, Peking University, Beijing 100871, China, and also with the Nat’l Eng. Research Center of Visual Technology, Beijing, 100871, China
        (e-mail \tt\small yizhou.wang@pku.edu.cn)}%
        \thanks{Project page: \href{https://sites.google.com/view/sasavan/}{https://sites.google.com/view/sasavan/}}
\thanks{Code available at: \href{https://github.com/wwwwwyyyyyxxxxx/SA2GVAN}{https://github.com/wwwwwyyyyyxxxxx/SA2GVAN}}
\thanks{Digital Object Identifier (DOI): 10.1109/LRA.2023.3272518.}
}%

\markboth{ IEEE ROBOTICS AND AUTOMATION LETTERS. PREPRINT VERSION. ACCEPTED APRIL 2023}
{Wang \MakeLowercase{\textit{et al.}}:Learning Semantic-Agnostic and Spatial-Aware Representation for Generalizable Visual-Audio Navigation}

\maketitle


\begin{abstract}
    Visual-audio navigation (VAN) is attracting more and more attention from the robotic community due to its broad applications, \emph{e.g.}, household robots and rescue robots.
    In this task, an embodied agent must search for and navigate to the sound source with egocentric visual and audio observations.
    However, the existing methods are limited in two aspects: 1) poor generalization to unheard sound categories; 2) sample inefficient in training.
    Focusing on these two problems, we propose a brain-inspired plug-and-play method to learn a semantic-agnostic and spatial-aware representation for generalizable visual-audio navigation.
    We meticulously design two auxiliary tasks for respectively accelerating learning representations with the above-desired characteristics.
    {With these two auxiliary tasks, the agent learns a spatially-correlated representation of visual and audio inputs that can be applied to work on environments with novel sounds and maps.}
    Experiment results on realistic 3D scenes (Replica and Matterport3D) demonstrate that our method achieves better generalization performance when zero-shot transferred to scenes with unseen maps and unheard sound categories.
\end{abstract}


\section{Introduction}
\label{intro}
\IEEEPARstart{E}{mbodied} agents should be able to navigate to different locations to complete downstream tasks such as goal-specific tidying 
and delivering items. Most robot navigation is currently limited to pure visual input from scenes~\cite{zhu2017target,DBLP:journals/corr/abs-1901-10915,DBLP:journals/corr/abs-1911-00357,objectgoalexp,https://doi.org/10.48550/arxiv.2203.10421}.
From a bionic perspective~\cite{LIU20097,guo2015audiovisual,gori2012development}, we humans can integrate audio information with visual observations to improve the ability to perceive objects and scenes, such as locating the position of an invisible object~\cite{carlson2013foundations}. Consequently,
it is advisable for an intelligent agent to learn how to perceive and leverage multi-modal information, including vision and audio, to achieve better navigation performance.

\begin{figure}[!h]
    \centering
    \vspace{-3.5mm}
    \includegraphics[width=0.46\textwidth,trim=0 00 0 0,clip]{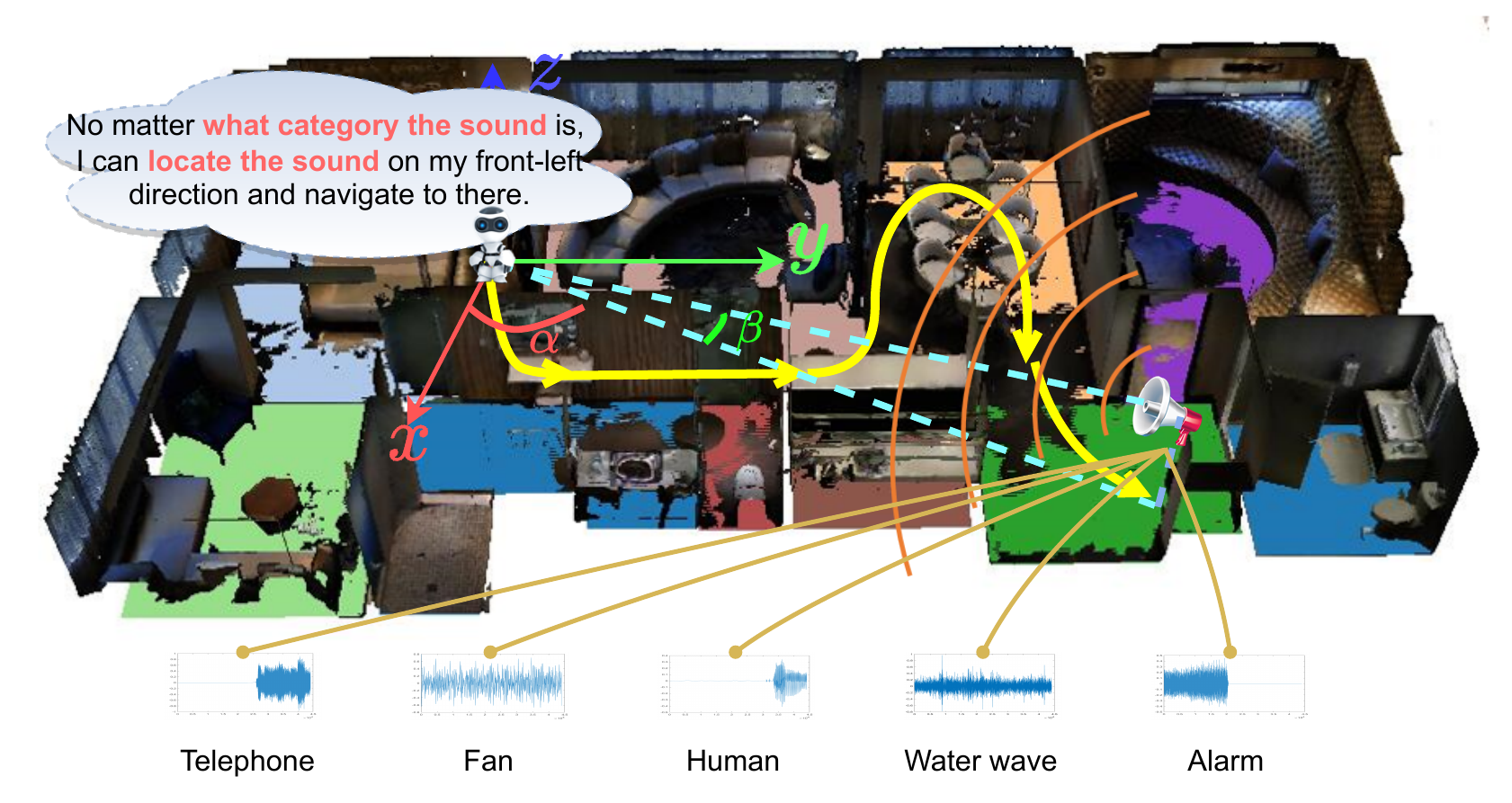}
    \caption{\textbf{Problem Setting.} The robot should navigate to the sound source location with the visual-audio observation, no matter what category of sound is being played. 
    In this example, the agent is in the bedroom initially and locates the sound in its front-left direction. {$\alpha$ and $\beta$ are the yaw and pitch angles of the sound source relative to the agent.}}
    \label{robot}
\end{figure}

With the recent development of the Soundspaces~\cite{chen_soundspaces_2020} simulation environment, 
researchers have begun to study leveraging both audio and visual information for navigation~\cite{chen_soundspaces_2020, chen_learning_2021, gan_look_2020}. 
In visual-audio navigation (VAN) task, testing sets include heard and unheard sound categories: 1) \emph{heard sound categories} mean the same sound categories group with the training set, 2) and \emph{unheard sound categories} are never heard sound categories by the agent during the training procedure. For training sets, in some specific and typical scenes, we can provide almost any kind of sound that might be present in these scenes.
For example, in a restaurant, a service robot may only need to learn to listen to service bells and customer greetings.
However, in some atypical and complex scenes, we cannot provide all possible sound categories to learn because of the wide range of sounds that the agent will confront, such as a guard robot that should
be able to react to odd sounds, activate the guard procedure and find where the odd sounds occur.
Therefore, intelligent agents need to handle unheard sound categories.
\IEEEpubidadjcol
Even though the state-of-the-art (SOTA) methods attain $\sim90\%$ success rate{~\cite{chen_soundspaces_2020,chen_learning_2021}} in Replica environments~\cite{replica} with heard sound categories, their success rates  drop to $\sim50\%$ when navigating to unheard sound.
Besides, existing methods use pure reinforcement learning loss (\emph{e.g.} critic loss and actor loss) to train an agent in a simulator and thus need about 3M$\sim$13M steps to converge due to low sample efficiency, which takes several days. It is important to develop an algorithm with high sample efficiency for this task.

Humans are sensitive to sounds, and even infants who know nothing about sound categories can perceive the general orientation of sound~\cite{GRAVEN2008187}. Motivated by the previous observation, in this paper, we refer to the human auditory processing mechanism. A dual-pathway model of auditory processing exists in the human brain where sound semantic information (\emph{what} path) and sound spatial information (\emph{where} path) are segregated into different brain areas~\cite{alain2008contribution,arnott2004assessing,yost1991auditory,heffner1990role}. 
Semantic information contains sound category and other category-related information, such as the percussive feeling of metal~\cite{adriani2003sound}. Spatial information includes the distance and direction of sounds
and other location-related information, such as the phase difference between two ears~\cite{arnott2004assessing,zundorf2016testing}. 
Semantic information changes with the sound category, 
leading to difficulties in learning generalizable semantic representations of unheard sound categories. In contrast, spatial information does not change~\cite{AcousticFields,cao2016interactive,veach1995bidirectional}, enabling the potential for generalizing to unheard sound categories. As a result, we opt to maintain different attention levels to different information in the features, \emph{i.e.}, to neglect semantic information and enhance spatial information. 

Concretely, based on the human auditory mechanism, we propose a plug-and-play method encouraging agents to learn task-relevant representations from multi-modal inputs. To improve sample efficiency and generalization in the VAN task, we design two auxiliary tasks that provide additional training signals. {These two tasks enable the agent to discover the intrinsic spatial correlations between visual and audio inputs. That can make it possible to apply the learned representation to environments with unseen sounds and maps.} In one auxiliary task, we use a gradient reversal layer to create an adversarial relationship between an audio encoder and an audio classifier to ignore semantic information. In the other auxiliary task, we use temporal information from visual and auditory inputs to predict the relative direction of a sound, thereby enhancing spatial information. Because our method is plug-and-play, it can be applied to various VAN backbone algorithms using the same settings. In our experiments, we use two SOTA algorithms, AV-Nav~\cite{chen_soundspaces_2020} and AV-Wan~\cite{chen_learning_2021} as the backbones.
We demonstrate the superiority of our proposed method on two realistic 3D scene datasets, Replica~\cite{replica} and Matterport3D~\cite{Matterport3D}, with strong generalization to scenarios with unheard sound categories and fewer training steps. In summary, our contributions are listed as follows:
\begin{enumerate}
    \item We observe that paying different attention to semantic and spatial components in sounds can improve the sample efficiency and the generalization of visual-audio navigators on unheard sound categories.

    \item We meticulously design two auxiliary tasks. One task uses an adversarial mechanism to neglect semantic information, and the other task predicts a relative direction to enhance spatial information.

    \item The experiments on two sets of realistic 3D scenes, Replica and Matterport3D, show that our method can achieve better generalization performance in fewer training steps.

\end{enumerate}

\section{Related Work}
\label{sec:related}   
\textbf{Visual-Audio Navigation.}
In this task, an agent should navigate to the sound source by utilizing egocentric visual and audio observations. 
The task is challenging because of the complexity of the room structure itself and its effect on sound propagation, which leads to the fact that the agent cannot precisely estimate the loudness and direction of the sound to make decisions.
Several existing studies~\cite{chen_soundspaces_2020,chen_learning_2021,gan_look_2020,chen2021semantic,tatiya2021knowledge} demonstrate the importance of fusing visual and audio modalities in navigation tasks and show good performance in scenes with heard sound categories. Some works~\cite{chen_soundspaces_2020,chen_learning_2021,gan_look_2020} do not explicitly focus on 
sound semantics and perform better on heard sound categories than unheard sound categories. Semantic-aware methods~\cite{chen2021semantic,tatiya2021knowledge} explicitly exploit the sound semantic information and learn the association between semantic information and scene representations to reason about the sound source location, \emph{e.g.}, hearing water dripping means the agent may need to go to the kitchen or bathroom.  
However, these semantic-aware methods~\cite{chen2021semantic,tatiya2021knowledge} can only deal with heard sound \textbf{categories}, including heard sound \textbf{instances} and unheard sound \textbf{instances}, while our method focuses on the generalization towards unheard sound \textbf{categories}. We argue that neglecting semantic information enhances the navigation generalization on unheard sound categories and does little harm or even improves the performance of heard sound categories.

\textbf{Auxiliary Task.}
It is not a new concept to train a reinforcement learning (RL) agent with auxiliary tasks. Auxiliary tasks are commonly used to improve the sample efficiency and attempt to build up state representations by predicting supplemental variables about important aspects of RL tasks, such as terminal state prediction~\cite{kartal2019terminal}, agent modeling~\cite{hernandez2019agent,foerster2017stabilising,hong2018deep}, return prediction~\cite{liu2020return,jaderberg2016reinforcement}, and depth prediction~\cite{mirowski2019learning}. Designing auxiliary tasks for a specific goal can be challenging, especially when the input contains multiple modalities. It is important to ensure consistency between the auxiliary tasks and the main task; otherwise, the auxiliary tasks will only train the agent to accomplish the auxiliary goals or hinder performance on the main task.
Our method introduces two auxiliary tasks for visual-audio navigation by referring to the human auditory mechanism. One is to predict the relative direction between the agent and the sound source location. Furthermore, the other is to force the agent to omit semantic information in sounds by adversarial learning.

\section{Method}
\label{sec:method}

\begin{figure*}[!t]
    \centering
    \includegraphics[width=1\textwidth
    ,trim=00 00 00 00,clip]{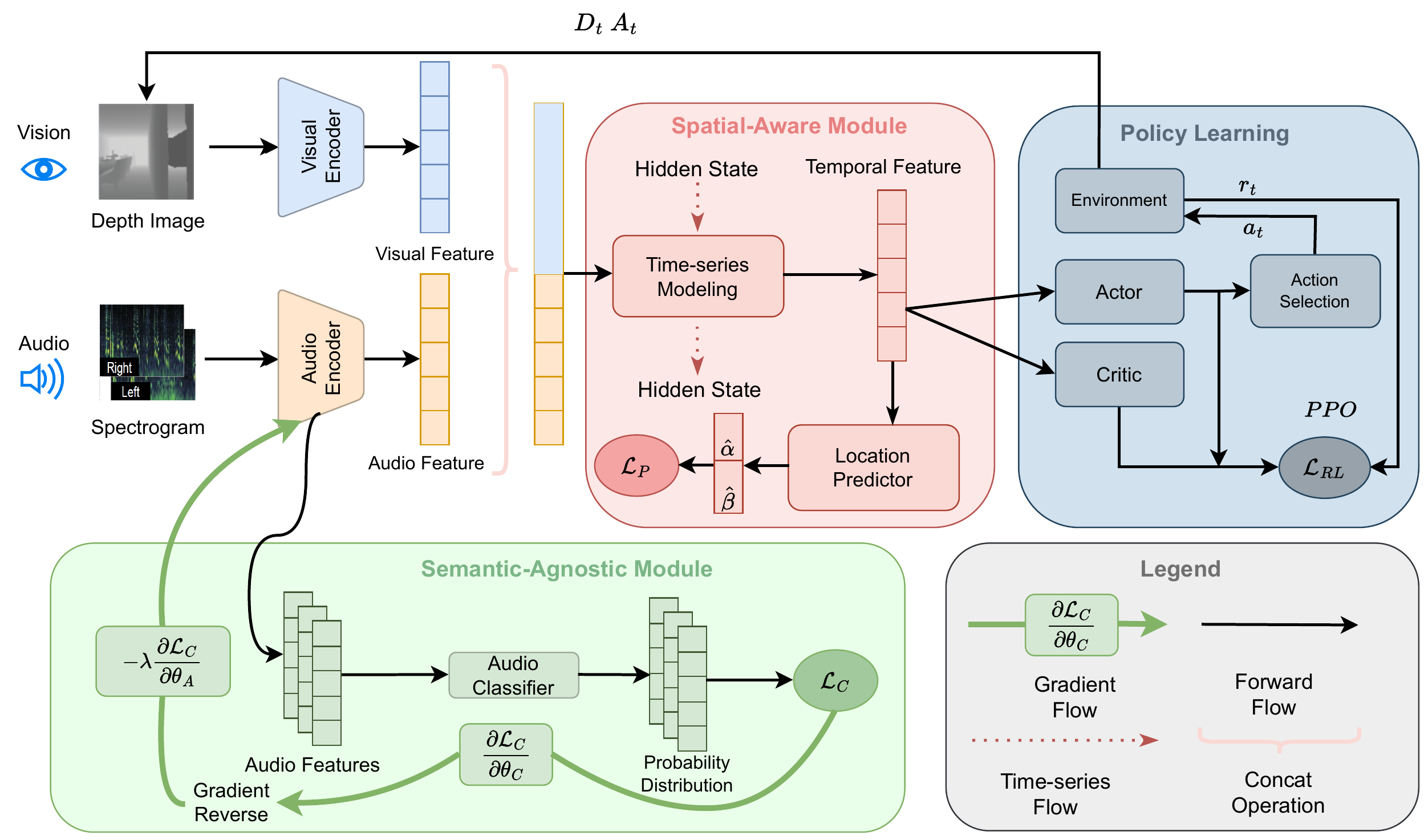}
    \caption{\textbf{Training Pipeline.} At each time step $t$, our method uses depth images ($D_t$) and spectrograms ($A_t$) as inputs for navigation. During the training procedure, an Audio Classifier (AC {, parameterized by $\theta_{C}$}) enforces the model to neglect semantic information via adversarial training supervised by $\mathcal L_C$. Concurrently, the temporal features ($O_t$) are given to a Location Predictor (LP) to pull out the sound source direction ($\alpha, \beta$) supervised by $\mathcal L_P$. {$\alpha$ and $\beta$ are the yaw and pitch angles of the sound source relative to the agent.} {Action Selection samples from the probability distribution generated by Actor to obtain action $a_t$. After executing $a_t$ in the environment, the environment returns a reward signal $r_t$.}
    At the end of each RL epoch, we train the Audio Encoder {(parameterized by $\theta_{A}$)}, the Audio Classifier and the Location Predictor simultaneously.}
    \label{art}
\end{figure*}
We follow the basic settings in AV-Nav\cite{2020VisualNavigation} and AV-Wan\cite{chen_learning_2021} for the AudioGoal Navigation task.
The task initializes an agent in the environment (a scene with single or multiple rooms) without the map of the environment. 
{In each episode, }a sound source is set in the environment, continuously emitting sounds that the agent can receive.
The agent is required to navigate to the sound source using visual and audio information. {All initial settings for the episodes are pre-generated, including the agent's initial position, the location of the target sound, the category of sound, and the room used for navigation, in order to avoid overly simplistic episodes.} 

In order to improve sample efficiency and make the navigation policy generalizable to unheard sound categories, we focus on extracting the generalizable components of the sounds referring to the human auditory mechanism.
The contents of sounds contain two main components: semantic information and spatial information. When the sound source location and robot position remain constant, the semantic information changes with the sound category, but the spatial information remains the same. 
 Our method is therefore composed of two main tasks for learning the generalizable representation: 
1) Semantic-Agnostic Learning (denoted in green in Fig.~\ref{art}) learning semantic-agnostic representation by an adversarial mechanism between audio encoder and audio classifier, and 2) Spatial-Aware Learning (denoted in red in Fig.~\ref{art}) learning spatial-aware representation by predicting the angle of sound relative to the agent by using a temporal representation containing visual and auditory information.

{Since the initial settings for each episode are pre-generated rather than randomly selected at the beginning of the episode~\cite{chen_soundspaces_2020}, without the Semantic-Agnostic Learning, the navigation policy will implicitly memorize the sounds used in each training episode (\emph{i.e.} over-fitting on the training episodes), so its generalization will be weakened.} Without Spatial-Aware Learning, Semantic-Agnostic Learning may mistakenly neglect the spatial information, making it also ignored by the agent (the most extreme case is that the audio encoder will output the same features for any audio input).


The additional processing of representations by these two tasks allows the agent to learn task-relevant features much faster, thus improving sample efficiency.


\subsection{Semantic-Agnostic Learning}
\label{sec:semanticagnostic}

When receiving a sound, a human may not know what the sound category exactly is but can estimate the sound source location~\cite{makous1990two, middlebrooks1991s,PICINALI2014393}, and even an infant who knows nothing about the world can roughly localize the sound source~\cite{morrongiello1990recent,morrongiello1994sound}, which shows that spatial information alone is sufficient for humans to locate sounds. 
Inspired by the research above, we argue that in AudioGoal navigation tasks for intelligent agents, spatial information of the sound is enough for locating and perceiving the sound. While semantic information changes with the sound categories, {it increases the difficulty for agents to learn generalizable semantic representations}. Moreover, for some atypical scenes (\emph{e.g.}, guard robots facing odd sounds), sounds and scenes are not closely related.
Therefore, learning semantic-agnostic representations should not harm the navigation performance on both heard sound categories but could enhance the generalization of unheard sound categories. 

Concretely, learning semantic-agnostic representations means that, with an agent fixed in a 
certain location and sound source in another certain location, the method outputs the same representation when taking sounds with different semantics.
To equip the representations learned by the method with the semantic-agnostic property, we design an auxiliary task in which 
an audio encoder needs to 
weaken the ability of the audio classifier to 
distinguish the current sound semantic category while the audio classifier attempts to distinguish the sound semantic category corresponding to an audio feature. The adversarial training forces the audio encoder to learn semantic-irrelevant representations.

Therefore, we use an adversarial mechanism between an audio encoder parameterized by $\theta_A$ and a 4-layer fully connected network audio classifier (AC) parameterized by $\theta_C$. To implement this adversarial mechanism, we employ a gradient reversal layer~\cite{ganin2015unsupervised} between the audio classifier and the audio encoder by multiplying a factor $-\lambda$ on gradient flow reflecting the adversarial intensity:
\begin{equation} 
        \label{lambda}
        \lambda=\frac{2b}{1.0+e^{-10*\frac{n}{N}}}-b
\end{equation}
where $n$ denotes the number of currently completed episodes, $N$ denotes the number of total episodes and $b$ denotes the bound of the adversarial intensity.
And the parameters are optimized as follows:
    \begin{equation} 
        \label{gdc}
        \theta_C \longleftarrow  \theta_C - \mu \frac{\partial{\mathcal{L}_C}}{\partial{\theta_C}}
        \end{equation}
    \begin{equation} 
        \label{gda}
    \theta_A \longleftarrow  \theta_A - \mu(\frac{\partial\mathcal{L}_O}{\partial{\theta_A}}-\lambda \frac{\partial\mathcal{L}_C}{\partial{\theta_A}})
    \end{equation}
    where $\mu$ denotes the learning rate, $\mathcal{L}_C$ denotes Cross Entropy Loss, and $\mathcal{L}_O$ denotes other loss related to $\theta_A$ such as Actor and Critic Loss in reinforcement learning.
    

\subsection{Spatial-Aware Learning}
\label{sec:predictor}
Semantic-agnostic learning ignores navigation-irrelevant information but does not encourage the agent to learn navigation-relevant representations. Although reinforcement learning provides reward signals to help the agent extract navigation-relevant features,  during the initial exploration phase, the agent may not catch sight of reward signals but {can rapidly learn} neglecting the semantic information of the sound from the adversarial audio classifier to minimize the adversarial optimization objective.
This rapid learning could lead to the audio encoder incorrectly ignoring spatial information as well, resulting in its output being insensitive to changes in the agent's position. On this occasion, the agent cannot navigate to the sound source.
Predicting sound location as an auxiliary task can effectively provide an additional training signal to help the agent extract spatial information and assist in navigation policy learning.

 We use a 4-layer fully connected network as the location predictor (LP) with temporal features generated by a Time-series Model as input to predict {the pitch and yaw angles of the sound source relative to the agent}, denoted as $\beta$ and $\alpha$ in Fig.~\ref{robot}, respectively. In practice, we do not predict the angle directly but predict the \emph{sine} and \emph{cosine} of the angle. The \emph{sine} and \emph{cosine} predictions avoid the periodicity of the angle that leads to the non-uniqueness. We use the Mean-Squared Loss as the auxiliary loss function. 
 {The gradients generated by the loss of the LP are utilized to update the Audio Encoder, the Visual Encoder, and the Time-series Model. These models can thus learn to extract features containing spatial information for RL's actor and critic to learn navigation policy better.}

\subsection{Training Details}
\label{traindetails}
We use SoundSpaces~\cite{chen_soundspaces_2020} as our simulator, enabling realistic audio rendering.
The SoundSpaces simulator discretizes scenes into uniformly distributed navigability graphs so that the agent can only move one node to a {naviagble} neighboring node in the graphs. {Where there are obstacles there are no nodes}. Thus the action space $\mathcal{A}$ has only four actions: $\mathrm{MoveForward}$, $\mathrm{TurnLeft}$, $\mathrm{TurnRight}$ and $\mathrm{Stop}$. {The Soundspace removes episodes where the distance from the start position to the target position is less than 4m and episodes where the shortest path is almost a straight line (ratio of geodesic to Euclidean distance less than 1.1).}

Since we apply our method on AV-Nav~\cite{chen_soundspaces_2020} and AV-Wan~\cite{chen_learning_2021}, we follow the design of their reward function, in which the agent is given a $+10$ reward if the agent executes action $Stop$ at the sound source location, $+1$ reward on AV-Nav or $+0.25$ reward on AV-Wan if the agent reduces the geodesic distance to the sound source location and an equivalent penalty if the agent increases the geodesic distance and $-0.01$ for time penalty.
 
 We train all learnable models jointly with Proximal Policy Optimization (PPO)~\cite{schulman2017proximal}. Each episode contains 150 steps, and the success criterion is met if the agent executes the action $\mathrm{Stop}$ at the sound position in 150 steps. 

\section{Experimental Results}
\label{sec:result}
\subsection{Experiment Settings}
\label{sec:expsetting}
\begin{table*}[!ht]
        \centering
        \caption{\textbf{Testing results on heard and unheard sound categories}. We apply our method to AV-Nav and AV-Wan and obtain higher quantitative results. The SPL and SNA show that our method improves the efficiency of the {previous works}, allowing the agent to choose a shorter path (higher SPL) and a faster path (higher SNA) to reach the sound location.}
        \label{tab:qresults}
\begin{tabular}{l|ccclll|llllll}
\hline
\multirow{3}{*}{Method} & \multicolumn{6}{c|}{Replica}                                                                                                                   & \multicolumn{6}{c}{MP3D}                                                                                                                  \\
                        & \multicolumn{3}{c}{Heard Sound Category}                                                    & \multicolumn{3}{c|}{Unheard Sound Category}      & \multicolumn{3}{c}{Heard Sound Category}                                               & \multicolumn{3}{c}{Unheard Sound Category}       \\ \cline{2-13} 
                        & SR($\uparrow$)                        & SPL($\uparrow$)                       & \multicolumn{1}{c|}{SNA($\uparrow$)}            & SR($\uparrow$)             & SPL($\uparrow$)            & SNA($\uparrow$)            & \multicolumn{1}{c}{SR($\uparrow$)} & \multicolumn{1}{c}{SPL($\uparrow$)} & \multicolumn{1}{c|}{SNA($\uparrow$)}            & SR($\uparrow$)             & SPL($\uparrow$)            & SNA($\uparrow$)            \\ \hline
Random                &  0.185 & 0.049 & \multicolumn{1}{c|}{0.018}          & 0.185          & 0.049          & 0.018          & 0.091                  & 0.021                   & \multicolumn{1}{c|}{0.008}          & 0.091          & 0.021          & 0.008          \\
DF                    &  0.720  & 0.547 & \multicolumn{1}{c|}{0.411}          & 0.111          & 0.172          & 0.084          & 0.412                  & 0.232                   & \multicolumn{1}{l|}{0.238}          & 0.180          & 0.139          & 0.107          \\ \hline
AV-Nav                & \textbf{0.903}                      & 0.672                     & \multicolumn{1}{c|}{0.354}          & 0.576          & 0.394          & 0.180          & \textbf{0.700}                  &\textbf{0.227}                   & \multicolumn{1}{l|}{\textbf{0.252}} & 0.359          & 0.105          & 0.109          \\
Ours+AV-Nav           & {0.898}            & \textbf{0.689}            & \multicolumn{1}{c|}{\textbf{0.400}} & \textbf{0.823} & \textbf{0.606} & \textbf{0.330} & {0.652}         & \textbf{0.227}          & \multicolumn{1}{l|}{0.217}          & \textbf{0.602} & \textbf{0.198} & \textbf{0.171} \\ \hline
AV-Wan                & \textbf{0.918}            & 0.676                     & \multicolumn{1}{c|}{0.522}          & 0.424          & 0.289          & 0.218          & 0.796                  & 0.453                   & \multicolumn{1}{l|}{0.337}          & 0.567          & 0.409          & 0.306          \\
Ours+AV-Wan           & 0.904                     & \textbf{0.709}            & \multicolumn{1}{c|}{\textbf{0.552}} & \textbf{0.628} & \textbf{0.434} & \textbf{0.330} & \textbf{0.829}         & \textbf{0.614}          & \multicolumn{1}{l|}{\textbf{0.468}} & \textbf{0.607} & \textbf{0.423} & \textbf{0.314} \\ \hline

\end{tabular}
        
\end{table*}
\begin{figure*}[!t]
\centering    
\subfloat[Replica AV-Nav SPL]
{
\includegraphics[width=0.235\textwidth]{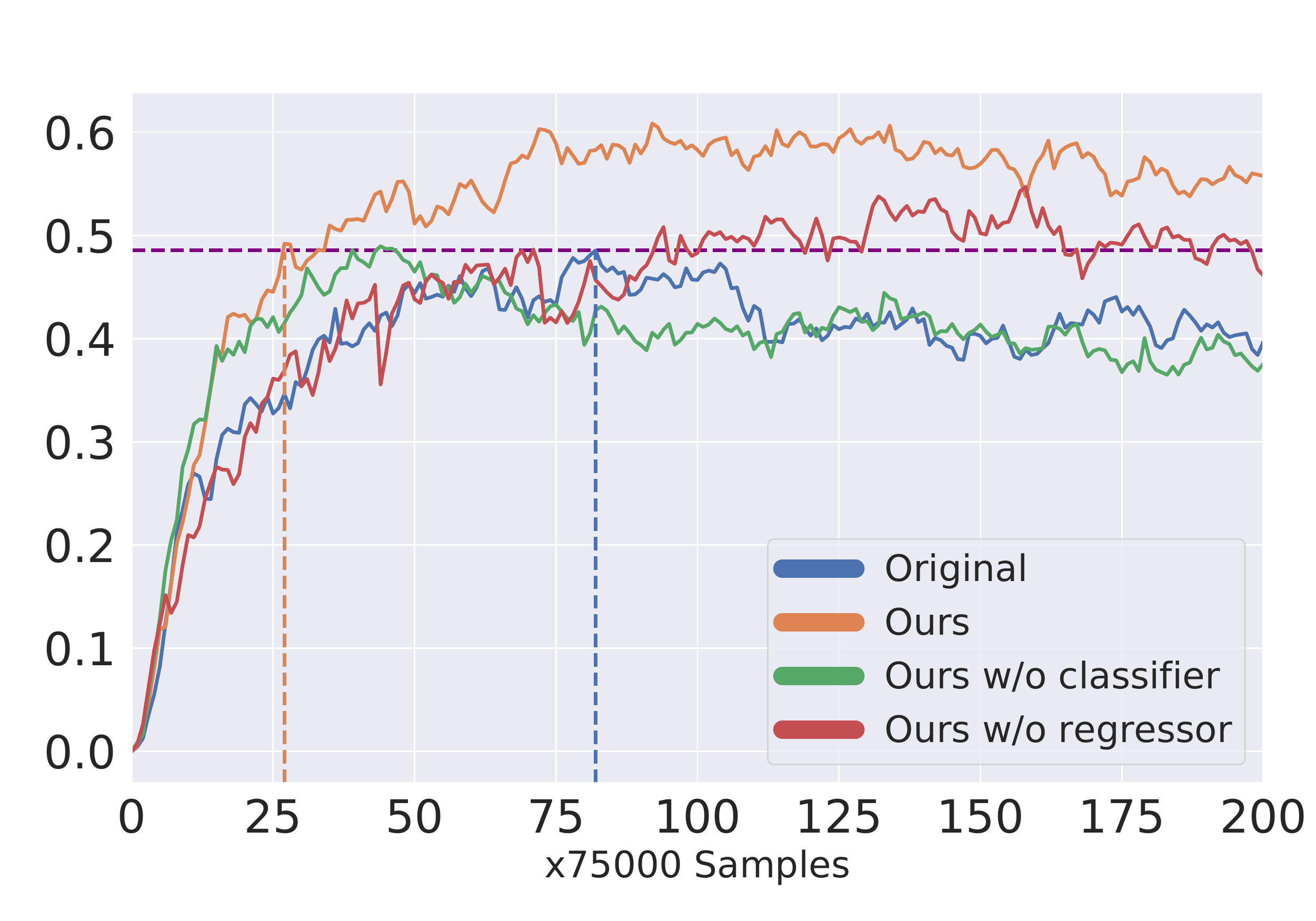}  
}     
\subfloat[Replica AV-Wan SPL]
{ 
\includegraphics[width=0.235\textwidth]{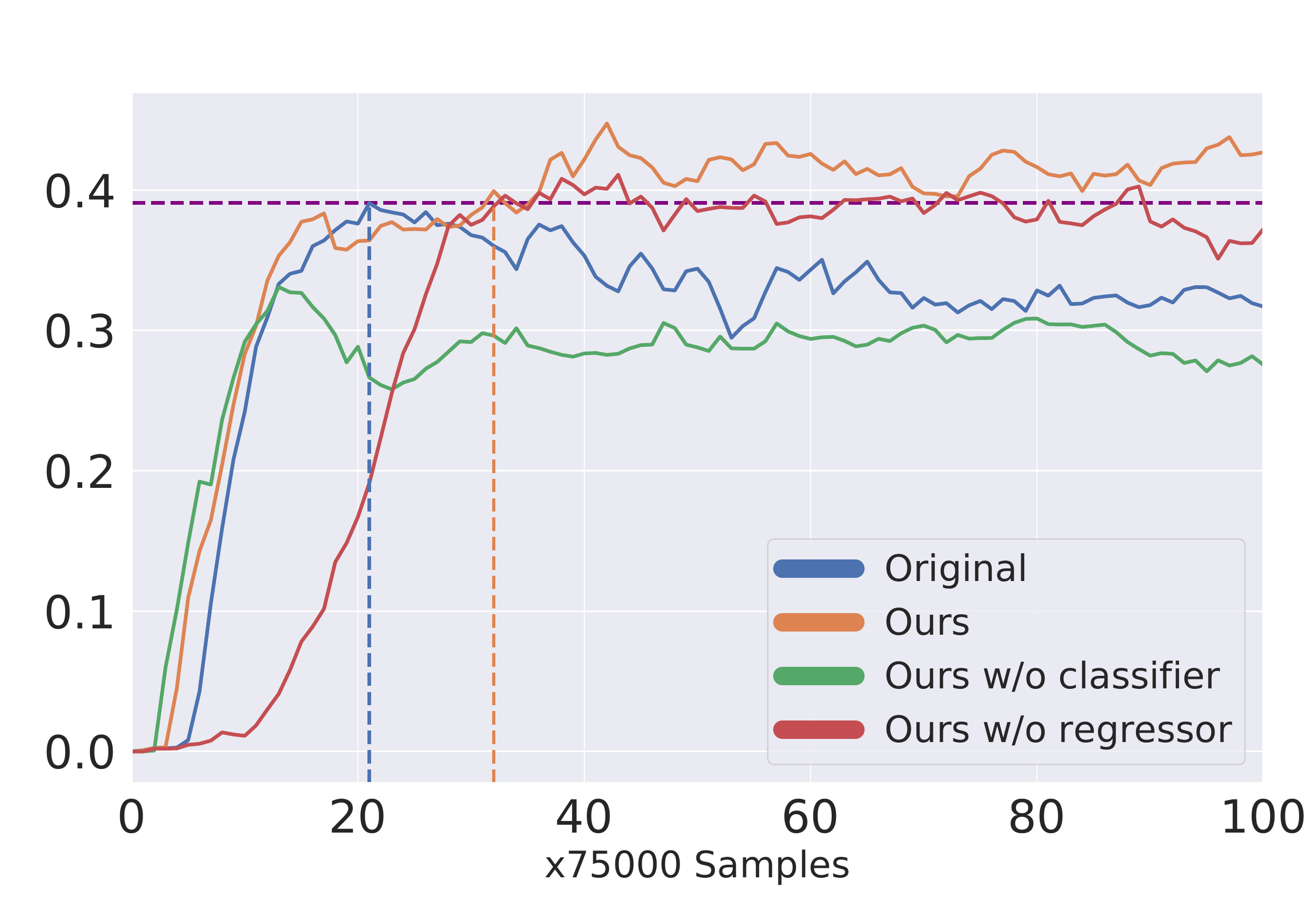}  
}     
\subfloat[MP3D AV-Nav SPL]
{ 
\includegraphics[width=0.235\textwidth]{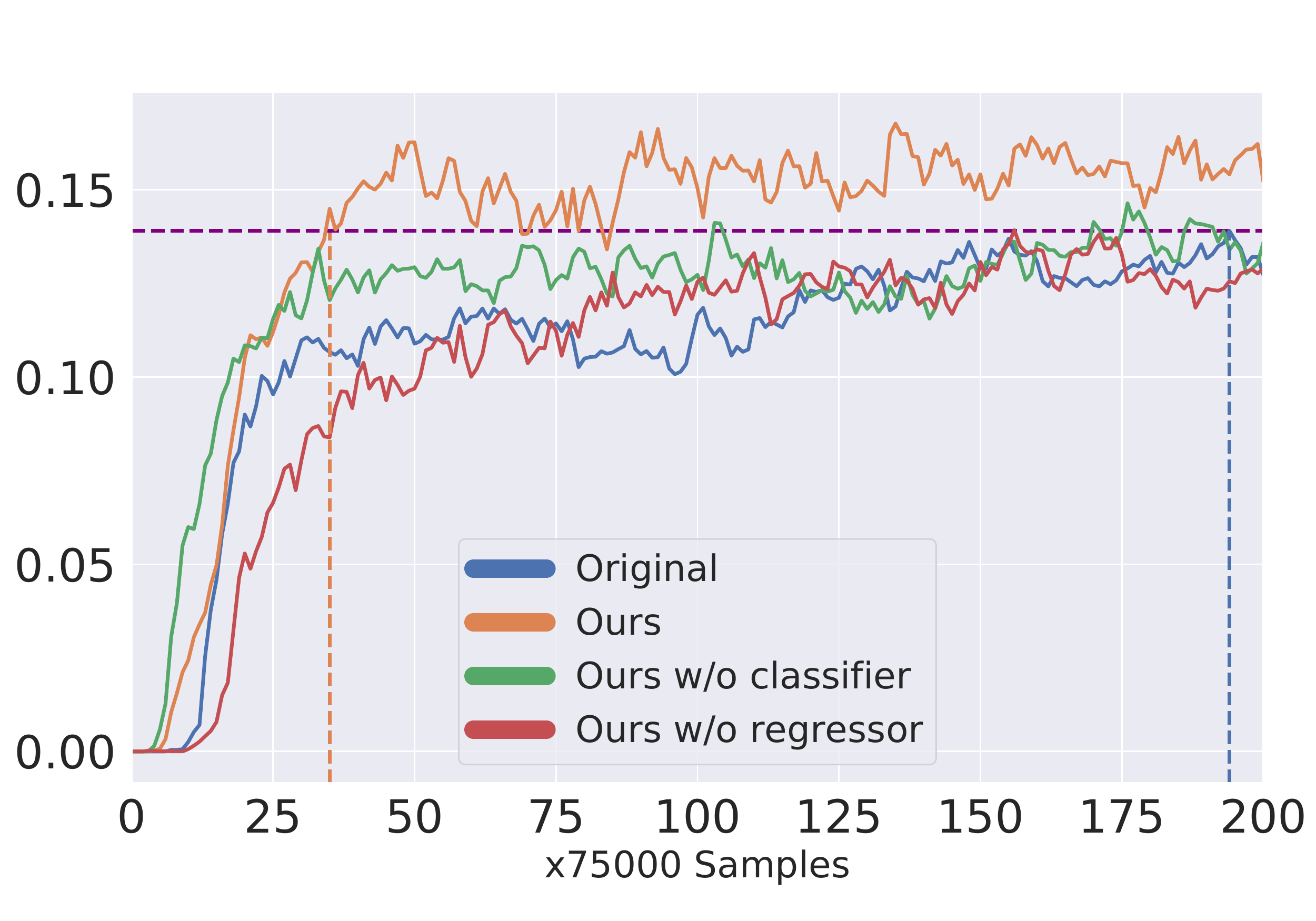}  
}     
\subfloat[MP3D AV-Wan SPL]
{ 
\includegraphics[width=0.235\textwidth]{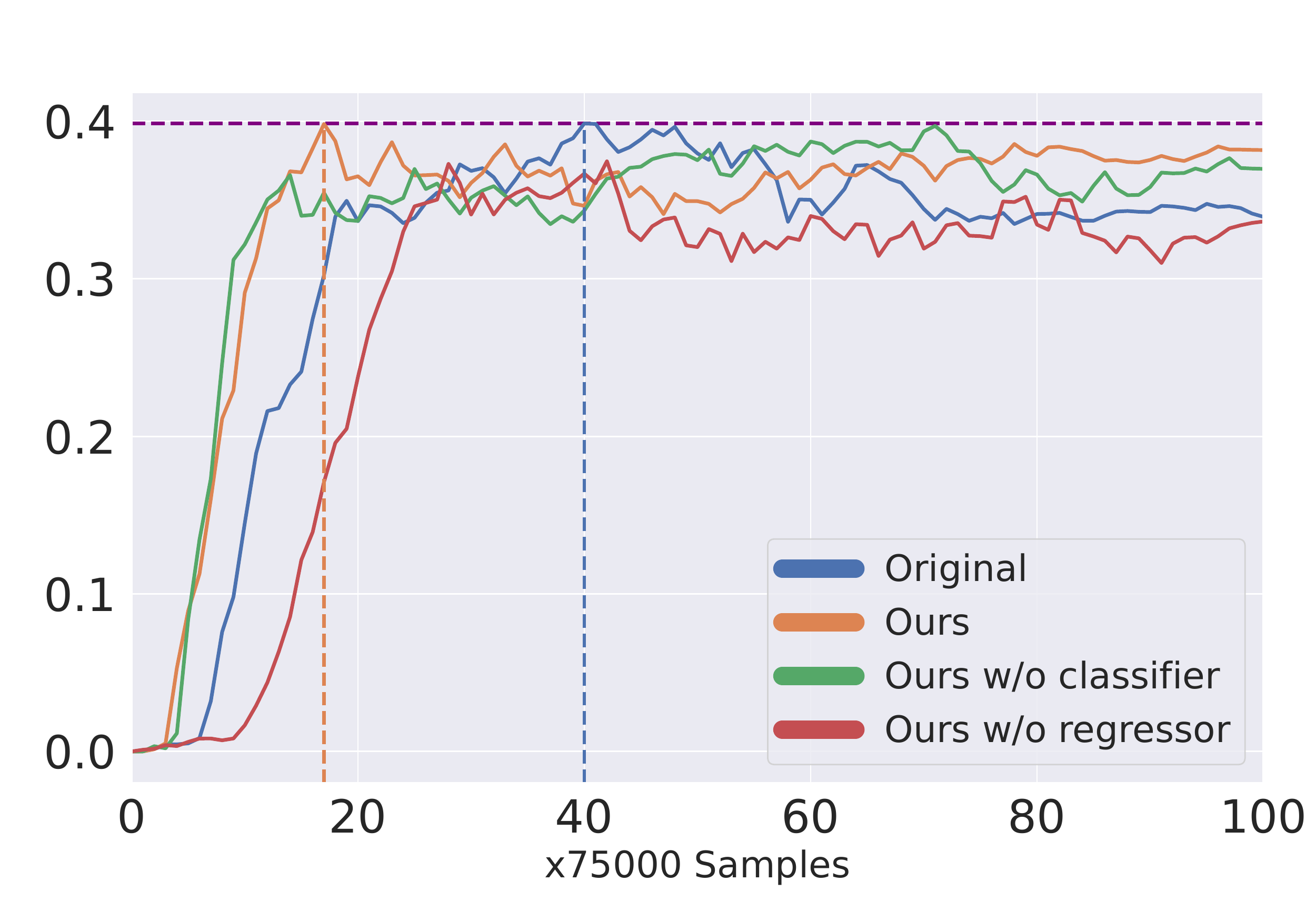}  
}     
\caption{\textbf{Learning Curve on testing sets.} We plot the testing results of the {previous works} and ours during training in both Replica and MatterPort3D environments with AV-Nav and AV-Wan as backbones, respectively. We plot a horizon dashed purple line across the highest SPL value of the {previous works} as a benchmark. We also draw vertical dashed lines for the {previous works} and ours in their corresponding colors, to indicate where their SPL values are greater than or equal to the benchmark for the first time. Our method can outperform the previous works with fewer training samples.}
\label{fig:curve}
\end{figure*}

\textbf{Environments and Datasets.} {We use the same audio and visual dataset and train/val/test splits as AV-Nav~\cite{chen_soundspaces_2020} and AV-Wan~\cite{chen_learning_2021} to demonstrate the improvement of our method.} We use the same simulator, SoundSpaces~\cite{chen_soundspaces_2020}, with two real-world 3D scene datasets, Replica and Matterport3D (MP3D), for training and testing our method along with train/val/test splits of 73/11/18 sound categories. 
Replica is a relatively small scene dataset with an average area of $47.24 m^2$ and train/val/test splits of 9/4/5 scenes. Matterport3D has relatively large scenes with an average area of $517.34 m^2$ and train/val/test splits of 57/10/12 scenes.
We also follow basic configuration and hyper-parameters from AV-Nav and AV-Wan {and only use depth maps as visual information}. 

\textbf{Metrics.} We evaluate our method on the following metrics: 
\begin{enumerate}
    \item Success Rate (SR): the fraction of successful episodes.
    \item Success Weighted by Path Length (SPL)~\cite{anderson2018evaluation}: we weigh the success by the ratio of the execution path length to the shortest path length.
    \item Success Weighted by Number of Actions (SNA){~\cite{chen_learning_2021}}: we weigh the success by the ratio of the executive action numbers to the minor action numbers.
\end{enumerate}
We use the model with the highest SPL on the validation set for testing and reporting the table results.

\textbf{Baselines.} We compare our methods with the following baselines:
\begin{enumerate}
    \item \textbf{Random}: an agent randomly selects an action in action space $\mathcal{A}$. The episode ends when executing $\mathrm{Stop}$.
    \item \textbf{Direction Follower(DF)}~\cite{chen_learning_2021}: This method pretrains a model to predict the direction of arrival (DoA). { An} agent sets an intermediate goal $K$ meters away in the predicted direction and plans to navigate there. We set $K=2$ in Replica and  $K=4$ in Matterport3D.
    \item \textbf{AV-Nav}~\cite{chen_soundspaces_2020}: it is a state-of-the-art VAN method that makes decisions using visual-audio fusion features with temporal sequences.
    \item \textbf{AV-Wan}~\cite{chen_learning_2021}: it is a state-of-the-art VAN method that builds geometric and acoustic maps and uses them to predict an intermediate goal adaptively. AV-Wan uses the Dijkstra~\cite{dijkstra_note_1959} shortest path algorithm to compute the path from the current node to the intermediate goal.
\end{enumerate}



\begin{figure*}[t]
\centering
    \includegraphics[width=1\textwidth
    ,trim=60 80 60 80,clip]{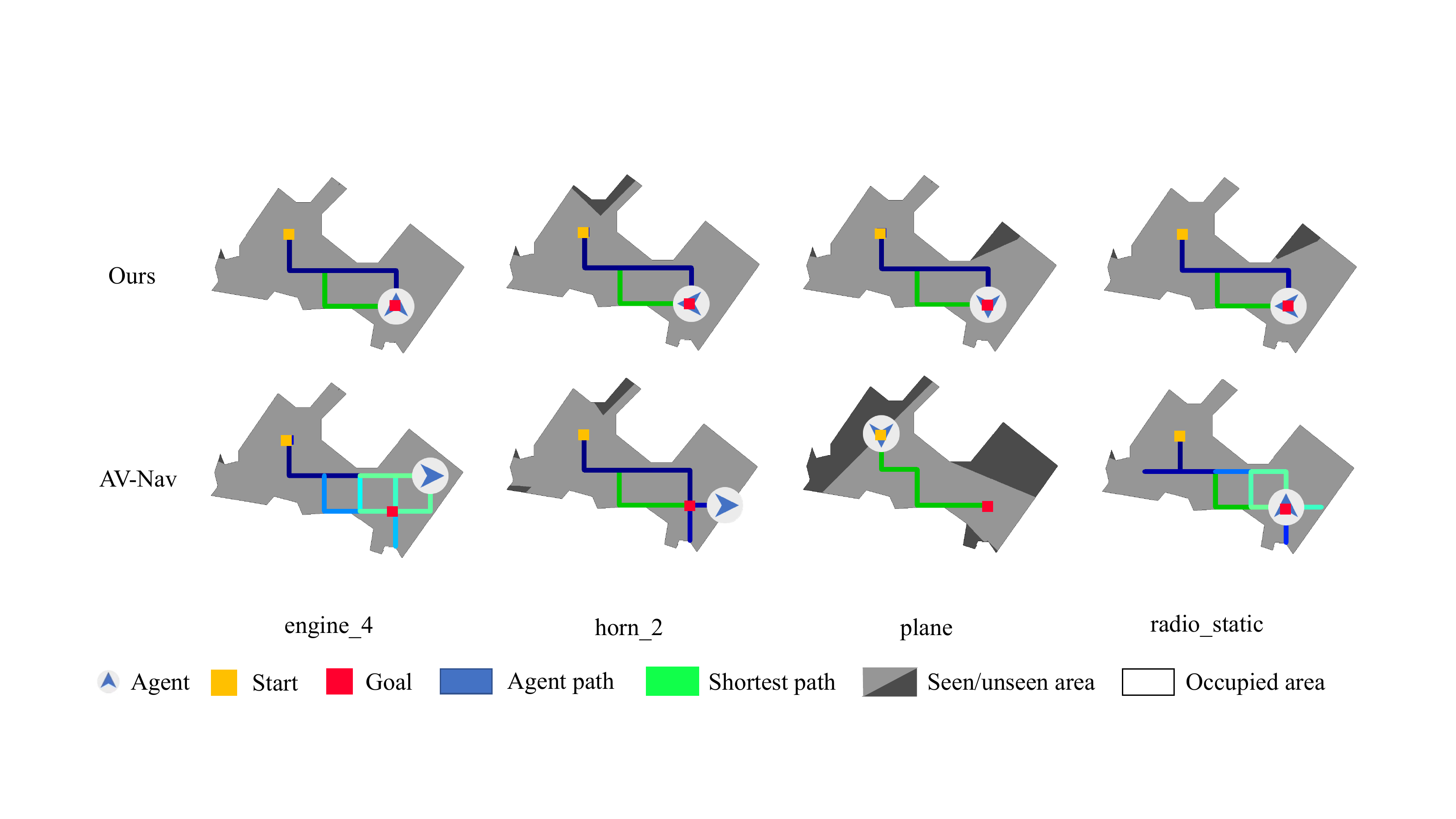}
    \caption{\textbf{Trajectory Visualization for different sound categories.} We visualized agent trajectories using our method and AV-Nav, respectively with the same set of start and end position {episodes} in the same scene. In each episode, the agent needs to navigate from the yellow point to the red point. The name at the bottom represents the category of sound, which means that each column has a different sound.  Agent path fades from dark blue to light blue as time goes by. Green is the shortest geodesic path in continuous space. { We aim to show that our method yields the same trajectory for different sound categories, which shows that the features we learn are indeed semantic-agnostic.}  The first row shows our results, and the second row is the results from AV-Nav. AV-Nav may fail in some {episodes}, \emph{e.g.}, the first three columns, and run quite differently when navigating to different sounds, while our method navigates to the goal in all four {episodes} and keep trajectory consistent in these {episodes}.}
    \label{fig:trajectory}
\end{figure*}

\subsection{Quantitative Comparison}

\begin{table}[!t]
\centering

\caption{\textbf{Ablation study for our method on AV-Nav.} We apply our method to AV-Nav and perform an ablation study on two components of our method, Audio Classifier (AC) and Location Predictor (LP), on testing sets of Replica and Matterport3D datasets. Three metrics are compared, including SR, SPL, and SNA.}
\vspace{5mm}
\label{tab:t3}
\begin{tabular}{cc|l|ccc}
\hline
                                     &  & \multicolumn{1}{c|}{Ablation} & SR($\uparrow$)             & SPL($\uparrow$)            & SNA($\uparrow$)            \\ \hline
{\multirow{8}{*}{\rotatebox{90}{Replica}}} & \multirow{4}{*}{\rotatebox{90}{AV-Nav}}       & Ours w/o AC and LP            & 0.576          & 0.394          & 0.180          \\
\multicolumn{1}{c}{}                         &                               & Ours w/o AC                   & 0.749          & 0.584          & 0.292          \\
\multicolumn{1}{c}{}                         &                               & Ours w/o LP                   & 0.766          & 0.588          & 0.266          \\
\multicolumn{1}{c}{}                         &                               & Ours                          & \textbf{0.823} & \textbf{0.606} & \textbf{0.330} \\ \cline{3-6} 
\multicolumn{1}{c}{}                         & \multirow{4}{*}{\rotatebox{90}{AV-Wan}}       & Ours w/o AC and LP            & 0.424          & 0.289          & 0.218          \\
\multicolumn{1}{c}{}                         &                               & Ours w/o AC                   & 0.432          & 0.319          & 0.292          \\
\multicolumn{1}{c}{}                         &                               & Ours w/o LP                   & 0.606          & 0.349          & 0.256          \\
\multicolumn{1}{c}{}                         &                               & Ours                          & \textbf{0.628} & \textbf{0.434} & \textbf{0.330} \\ \hline
\multirow{8}{*}{\rotatebox{90}{MP3D}}                        & \multirow{4}{*}{\rotatebox{90}{AV-Nav}}       & Ours w/o AC and LP            & 0.359          & 0.105          & 0.109          \\
                                             &                               & Ours w/o AC                   & 0.561          & 0.169          & 0.154          \\
                                             &                               & Ours w/o LP                   & 0.478          & 0.143          & 0.150          \\
                                             &                               & Ours                          & \textbf{0.602} & \textbf{0.198} & \textbf{0.171} \\ \cline{3-6} 
                                             & \multirow{4}{*}{\rotatebox{90}{AV-Wan}}       & Ours w/o AC and LP            & 0.567          & 0.409          & 0.306          \\
                                             &                               & Ours w/o AC                   & 0.556          & 0.399          & 0.305          \\
                                             &                               & Ours w/o LP                   & 0.599          & 0.391          & 0.295          \\
                                             &                               & Ours                          & \textbf{0.607} & \textbf{0.423} & \textbf{0.314} \\ \hline
\end{tabular}
\end{table}
    \label{quanti}
   We apply our method on AV-Nav~\cite{chen_soundspaces_2020} and AV-Wan~\cite{chen_learning_2021} and test baselines and our method referred by Ours+AV-Nav and Ours+AV-Wan on unheard sound categories in Tab.~\ref{tab:qresults}.
   
   Random performs poorly on both datasets, showing that the difficulty of the task and the robot is supposed to make good use of visual and audio cues. Direction Follower uses only audio information for decision making, while visual information is only used for path planning, so Direction Follower performs worse than the method that fuses information from both modalities to make decisions. 

 %
   After applying our method, AV-Nav and AV-Wan achieve significant improvements on Replica and Matterport3D datasets on unheard sound categories, proving that our method works well for different backbone algorithms and datasets. In particular, on Replica, our method gains about 50\%  SPL improvement on the {previous works}.
   The results on AV-Nav and AV-Wan demonstrate the advantages of our method {where} we optimize the features and represent them in a more task-specific manner. We also test our method on heard sound categories, shown in Tab.~\ref{tab:qresults}. 
   The results show that our method improves performance slightly, showing that our method does not trade performance on the heard sound categories for generalizability by impairing it. 

\begin{table*}[!ht]
\centering
\caption{{\textbf{Audio Noise Experiments.} We show the SPL in the experiments with different levels of noise(following the noise settings from AV-Wan~\cite{chen_learning_2021}). Our method still outperforms the previous works in most cases, and the performance does not show a large degradation compared to the noise-free experiments, showing our method's robustness to noise. We present the average SPL on differenet noise levels in the additional columns.}}
\vspace{-0cm}
\resizebox{\linewidth}{!}{
\begin{tabular}{l|cccccccccc|cccccccccc}
\hline
                        & \multicolumn{10}{c|}{Replica}                                                                                                                                                                                                                               & \multicolumn{10}{c}{Mp3d}                                                                                                                                                                                                                                   \\ \hline
Audio Noise Level (SNR) & \multicolumn{2}{c|}{50}                              & \multicolumn{2}{c|}{40}                              & \multicolumn{2}{c|}{30}                              & \multicolumn{2}{c|}{20}                              & \multicolumn{2}{c|}{Avg. SPL}   & \multicolumn{2}{c|}{50}                              & \multicolumn{2}{c|}{40}                              & \multicolumn{2}{c|}{30}                              & \multicolumn{2}{c|}{20}                              & \multicolumn{2}{c}{Avg. SPL}    \\ \hline
Depth Noise             & w              & \multicolumn{1}{c|}{w/o}            & w              & \multicolumn{1}{c|}{w/o}            & w              & \multicolumn{1}{c|}{w/o}            & w              & \multicolumn{1}{c|}{w/o}            & w              & w/o            & w              & \multicolumn{1}{c|}{w/o}            & w              & \multicolumn{1}{c|}{w/o}            & w              & \multicolumn{1}{c|}{w/o}            & w              & \multicolumn{1}{c|}{w/o}            & w              & w/o            \\ \hline
AV-Nav                  & 0.327          & \multicolumn{1}{c|}{0.363}          & 0.360          & \multicolumn{1}{c|}{0.362}          & 0.266          & \multicolumn{1}{c|}{0.300}          & 0.243          & \multicolumn{1}{c|}{0.305}          & 0.299          & 0.333          & 0.102          & \multicolumn{1}{c|}{0.103}          & 0.101          & \multicolumn{1}{c|}{0.111}          & 0.104          & \multicolumn{1}{c|}{0.107}          & 0.118          & \multicolumn{1}{c|}{0.124}          & 0.106          & 0.111          \\
Ours+AV-Nav             & \textbf{0.398} & \multicolumn{1}{c|}{\textbf{0.401}} & \textbf{0.432} & \multicolumn{1}{c|}{\textbf{0.461}} & \textbf{0.452} & \multicolumn{1}{c|}{\textbf{0.472}} & \textbf{0.405} & \multicolumn{1}{c|}{\textbf{0.442}} & \textbf{0.422} & \textbf{0.444} & \textbf{0.114} & \multicolumn{1}{c|}{\textbf{0.143}} & \textbf{0.131} & \multicolumn{1}{c|}{\textbf{0.133}} & \textbf{0.149} & \multicolumn{1}{c|}{\textbf{0.146}} & \textbf{0.139} & \multicolumn{1}{c|}{\textbf{0.140}} & \textbf{0.133} & \textbf{0.141} \\ \hline
AV-Wan                  & 0.291          & \multicolumn{1}{c|}{0.275}          & 0.286          & \multicolumn{1}{c|}{0.275}          & \textbf{0.337} & \multicolumn{1}{c|}{\textbf{0.352}} & 0.314          & \multicolumn{1}{c|}{0.307}          & 0.307          & 0.302          & 0.279          & \multicolumn{1}{c|}{0.279}          & 0.314          & \multicolumn{1}{c|}{0.314}          & 0.331          & \multicolumn{1}{c|}{0.331}          & \textbf{0.375} & \multicolumn{1}{c|}{\textbf{0.375}} & 0.325          & 0.325          \\
Ours+AV-Wan             & \textbf{0.367} & \multicolumn{1}{c|}{\textbf{0.368}} & \textbf{0.354} & \multicolumn{1}{c|}{\textbf{0.353}} & 0.327          & \multicolumn{1}{c|}{0.338}          & \textbf{0.347} & \multicolumn{1}{c|}{\textbf{0.350}} & \textbf{0.349} & \textbf{0.352} & \textbf{0.297} & \multicolumn{1}{c|}{\textbf{0.297}} & \textbf{0.377} & \multicolumn{1}{c|}{\textbf{0.377}} & \textbf{0.342} & \multicolumn{1}{c|}{\textbf{0.342}} & 0.361          & \multicolumn{1}{c|}{0.361}          & \textbf{0.344} & \textbf{0.344} \\ \hline
\end{tabular}
}
\label{tab:noise}
\end{table*}
    {Considering that there exist domain gaps between the real world and the simulator, such as audio and depth noise, we add these two parts of noise to the environment to simulate the real world and demonstrate the robustness of our method following the setting of audio noise and depth noise from AV-Wan~\cite{chen_learning_2021}. We conducted experiments on noise levels ranging from 20 to 50, with intervals of 10. Notice that, while AV-Wan~\cite{chen_learning_2021} only use \emph{telephone} in the noise experiments as the target sound, our work focuses on the generalization ability towards unheard sound categories, so we use all the sound categories in the testing set as target sounds instead. The results are shown in Tab.~\ref{tab:noise}. Note that even with different noise levels, our method still improves the performance of the previous works. With different levels of noise, the performance of our method shows no significant degradation and exhibits strong robustness.
    The robustness to noise can indicate that our method has the potential to be used in the real world.}

\subsection{Sample Efficiency and Learning Curve}

To demonstrate our method's high sample efficiency, we show the learning curves on the testing set
on the Replica and MatterPort3D with both AV-Nav and AV-Wan as backbones. 
Fig.~\ref{fig:curve} shows that our method can achieve higher performance than the final results of the {previous works}, with fewer samples than the {previous works} needs to converge. We compare the number of samples required by ours and the {previous works}, using the highest point of the {previous works} as a benchmark. In Fig.~\ref{fig:curve} (a), (c), and (d), our methods require fewer samples, and the performance still grows as the samples grow. In Fig.~\ref{fig:curve} (b), although there is no significant sample difference between ours and the {previous works}, our method is more stable in the later stages and the performance continues to grow.

\subsection{Trajectory Visualizations}
\begin{figure*}[!ht]
\centering
    \includegraphics[width=1\textwidth
    ,trim=00 40 00 130,clip]{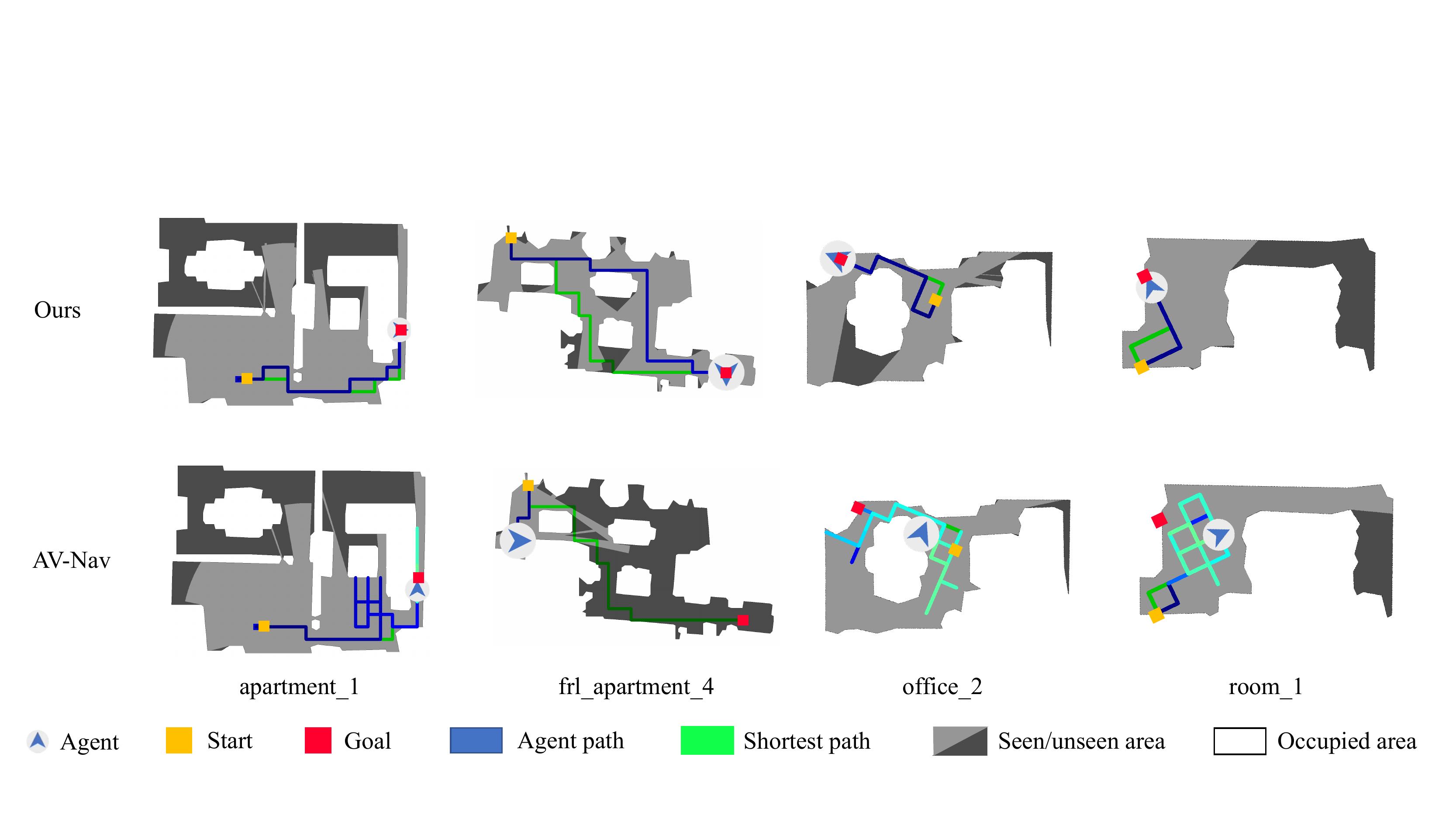}
    \caption{\textbf{Trajectory Visualization for Different Scenes.}We visualized navigation trajectories using our method and AV-Nav in various scenes. The name at the bottom represents the scene. {In each episode, the agent needs to navigate from the yellow point to the red point.} Agent path fades from dark blue to light blue as time goes by. Green is the shortest geodesic path in continuous space. The first row shows our results, and the second row is the results from AV-Nav. AV-Nav may fail in some {episodes}, \emph{e.g.}, the second and third column, or take a complex route, \emph{e.g.}, the first, third, and fourth column. {Our method finds a good path to the end point in all four episodes.}}
    \label{fig:trajectory_comparison}
    \vspace{-0.5cm}
\end{figure*}
\label{sec:trajvis}
We visualize the trajectories using our method and AV-Nav under four categories of sounds, shown in Fig.~\ref{fig:trajectory}.
We refer to the same start agent position and the same sound source location within the same scene as the \emph{same task}. 
To view the trajectory generation process, please watch the attached video.
In the first line of  Fig.~\ref{fig:trajectory}, our method can come out of the trajectory equivalent to the shortest path in various sounds consistent with each other. In the second line, however, AV-Nav either fails to complete the task or the trajectory is very complex and inconsistent.

We also visualize the trajectories in different scenes, shown in Fig.~\ref{fig:trajectory_comparison}. Our method can generate more efficient trajectories within different scenes than AV-Nav.




\subsection{Ablation Studies and Analysis}
Tab.~\ref{tab:t3} shows the ablation results of the audio classifier and the location predictor components of our method. Removing either the audio classifier or the location predictor leads to a reduction in performance. Notably, 
reducing the location predictor hurts the performance more than reducing the audio classifier does in Matterport3D. Compared to Replica, scenes in Matterport3D have bigger areas; thus, spatial information is more helpful in completing tasks in Matterport3D.

In addition, the audio classifier provides an adversarial training mechanism, which implicitly boosts the model's generalization by forcing the model to ignore the semantic information of the audio inputs. Meanwhile, the model can benefit from the auxiliary localization task's additional training signals and directly improve navigation performance.

\section{Conclusion and Discussion}
\label{sec:conclusion}
This work focuses on the generalization and sample efficiency problem for VAN tasks.
The different properties of spatial and semantic information inspired us to reduce the generalization gap between unheard and heard sound categories and learn task-relevant representations fast.
Therefore, we propose a plug-and-play method to narrow the performance gap on unheard and heard sound categories by neglecting semantic information while enhancing spatial information. 
Evaluations on Replica and Matterport3D show that our method significantly outperforms the baseline on the unheard sound categories and slightly improves the heard sound categories. Learning curves show that our method has better sample efficiency than baselines.
{We also conducted audio and depth noise experiments to demonstrate the robustness of our method to depth image noise and varying levels of audio noise. The results show that our method performs well even with noisy inputs.}

{
In the future, we will further explore the methods to enhance the generalization in more challenging visual-audio navigation settings, e.g., real-world development and complex environments.  
1) Real-world development (sim2real transfer) involves the challenging task of transferring reinforcement learning models trained in simulated environments to real robots. Due to the significant sim2real gap in both audio and visual modalities, conducting experiments in the real world remains difficult. To overcome this challenge, we must address the discrepancy between simulation and reality and improve the model's generalization ability. One potential solution is to apply bi-directional domain adaptation to align the feature distributions of simulation and reality during training. Additionally, exploring meta-reinforcement learning algorithms may enable the agent to efficiently mitigate domain drift during test time.
2) In complex environments, the agent must handle interference from multiple sound sources and uncertainty from moving sound.
To tackle scenarios with multiple sound sources at similar volume levels, we can leverage semantic information and sound source separation algorithms~\cite{majumder2022active, tzinis2020two} to filter out the target sound source as input to the navigator. Moreover, we can augment the training process with a multi-agent game~\cite{zhong2021towards} to automatically generate diverse and challenging distracting or moving sources, further enhancing the robustness of the system.
}

\section{Acknowledgement}
This project was supported by MOST-2022ZD0114900, NSFC No.62006006, No. 62136001, No.62061136001 and Qualcomm University Research Grant.

\bibliographystyle{IEEEtran}
\bibliography{example}

\begin{thebibliography}{10}
\providecommand{\url}[1]{#1}
\csname url@samestyle\endcsname
\providecommand{\newblock}{\relax}
\providecommand{\bibinfo}[2]{#2}
\providecommand{\BIBentrySTDinterwordspacing}{\spaceskip=0pt\relax}
\providecommand{\BIBentryALTinterwordstretchfactor}{4}
\providecommand{\BIBentryALTinterwordspacing}{\spaceskip=\fontdimen2\font plus
\BIBentryALTinterwordstretchfactor\fontdimen3\font minus
  \fontdimen4\font\relax}
\providecommand{\BIBforeignlanguage}[2]{{%
\expandafter\ifx\csname l@#1\endcsname\relax
\typeout{** WARNING: IEEEtran.bst: No hyphenation pattern has been}%
\typeout{** loaded for the language `#1'. Using the pattern for}%
\typeout{** the default language instead.}%
\else
\language=\csname l@#1\endcsname
\fi
#2}}
\providecommand{\BIBdecl}{\relax}
\BIBdecl

\bibitem{zhu2017target}
Y.~Zhu, R.~Mottaghi, E.~Kolve, J.~J. Lim, A.~Gupta, L.~Fei-Fei, and A.~Farhadi,
  ``Target-driven visual navigation in indoor scenes using deep reinforcement
  learning,'' in \emph{ICRA}.\hskip 1em plus 0.5em minus 0.4em\relax IEEE,
  2017, pp. 3357--3364.

\bibitem{DBLP:journals/corr/abs-1901-10915}
D.~Mishkin, A.~Dosovitskiy, and V.~Koltun, ``Benchmarking classic and learned
  navigation in complex 3d environments,'' \emph{CoRR}, vol. abs/1901.10915,
  2019.

\bibitem{DBLP:journals/corr/abs-1911-00357}
E.~Wijmans, A.~Kadian, A.~Morcos, S.~Lee, I.~Essa, D.~Parikh, M.~Savva, and
  D.~Batra, ``Decentralized distributed {PPO:} solving pointgoal navigation,''
  \emph{CoRR}, vol. abs/1911.00357, 2019.

\bibitem{objectgoalexp}
D.~S. Chaplot, D.~P. Gandhi, A.~Gupta, and R.~R. Salakhutdinov, ``Object goal
  navigation using goal-oriented semantic exploration,'' \emph{NIPS}, vol.~33,
  pp. 4247--4258, 2020.

\bibitem{https://doi.org/10.48550/arxiv.2203.10421}
S.~Y. Gadre, M.~Wortsman, G.~Ilharco, L.~Schmidt, and S.~Song, ``Clip on
  wheels: Zero-shot object navigation as object localization and exploration,''
  2022.

\bibitem{LIU20097}
B.~Liu, Z.~Wang, and Z.~Jin, ``The integration processing of the visual and
  auditory information in videos of real-world events: An erp study,''
  \emph{Neuroscience Letters}, vol. 461, no.~1, pp. 7--11, 2009.

\bibitem{guo2015audiovisual}
X.~Guo, X.~Li, X.~Ge, and S.~Tong, ``Audiovisual congruency and incongruency
  effects on auditory intensity discrimination,'' \emph{Neuroscience Letters},
  vol. 584, pp. 241--246, 2015.

\bibitem{gori2012development}
M.~Gori, G.~Sandini, and D.~Burr, ``Development of visuo-auditory integration
  in space and time,'' \emph{Frontiers in integrative neuroscience}, vol.~6,
  p.~77, 2012.

\bibitem{carlson2013foundations}
N.~R. Carlson, \emph{Foundations of behavioral neuroscience}.\hskip 1em plus
  0.5em minus 0.4em\relax Pearson Education, 2013.

\bibitem{chen_soundspaces_2020}
C.~Chen, U.~Jain, C.~Schissler, S.~V.~A. Gari, Z.~Al-Halah, V.~K. Ithapu,
  P.~Robinson, and K.~Grauman, ``Soundspaces: Audio-visual navigation in 3d
  environments,'' in \emph{ECCV}.\hskip 1em plus 0.5em minus 0.4em\relax
  Springer, 2020, pp. 17--36.

\bibitem{chen_learning_2021}
C.~Chen, S.~Majumder, Z.~Al-Halah, R.~Gao, S.~K. Ramakrishnan, and K.~Grauman,
  ``Learning to set waypoints for audio-visual navigation,'' in \emph{ICLR},
  2020.

\bibitem{gan_look_2020}
C.~Gan, Y.~Zhang, J.~Wu, B.~Gong, and J.~B. Tenenbaum, ``Look, listen, and act:
  Towards audio-visual embodied navigation,'' in \emph{ICRA}.\hskip 1em plus
  0.5em minus 0.4em\relax IEEE, 2020, pp. 9701--9707.

\bibitem{replica}
J.~Straub, T.~Whelan, L.~Ma, Y.~Chen, E.~Wijmans, S.~Green, J.~J. Engel,
  R.~Mur{-}Artal, C.~Ren, S.~Verma, A.~Clarkson, M.~Yan, B.~Budge, Y.~Yan,
  X.~Pan, J.~Yon, Y.~Zou, K.~Leon, N.~Carter, J.~Briales, T.~Gillingham,
  E.~Mueggler, L.~Pesqueira, M.~Savva, D.~Batra, H.~M. Strasdat, R.~D. Nardi,
  M.~Goesele, S.~Lovegrove, and R.~A. Newcombe, ``The replica dataset: {A}
  digital replica of indoor spaces,'' \emph{CoRR}, vol. abs/1906.05797, 2019.

\bibitem{GRAVEN2008187}
S.~N. Graven and J.~V. Browne, ``Auditory development in the fetus and
  infant,'' \emph{Newborn and Infant Nursing Reviews}, vol.~8, no.~4, pp.
  187--193, 2008, brain Development of the Neonate.

\bibitem{alain2008contribution}
C.~Alain, Y.~He, and C.~Grady, ``The contribution of the inferior parietal lobe
  to auditory spatial working memory,'' \emph{Journal of cognitive
  neuroscience}, vol.~20, no.~2, pp. 285--295, 2008.

\bibitem{arnott2004assessing}
S.~R. Arnott, M.~A. Binns, C.~L. Grady, and C.~Alain, ``Assessing the auditory
  dual-pathway model in humans,'' \emph{Neuroimage}, vol.~22, no.~1, pp.
  401--408, 2004.

\bibitem{yost1991auditory}
W.~A. Yost, ``Auditory image perception and analysis: The basis for hearing,''
  \emph{Hearing research}, vol.~56, no. 1-2, pp. 8--18, 1991.

\bibitem{heffner1990role}
H.~E. Heffner and R.~S. Heffner, ``Role of primate auditory cortex in
  hearing,'' \emph{Comparative perception}, vol.~2, 1990.

\bibitem{adriani2003sound}
M.~Adriani, P.~Maeder, R.~Meuli, A.~B. Thiran, R.~Frischknecht, J.-G.
  Villemure, J.~Mayer, J.-M. Annoni, J.~Bogousslavsky, E.~Fornari
  \emph{et~al.}, ``Sound recognition and localization in man: specialized
  cortical networks and effects of acute circumscribed lesions,''
  \emph{Experimental brain research}, vol. 153, no.~4, pp. 591--604, 2003.

\bibitem{zundorf2016testing}
I.~C. Z{\"u}ndorf, J.~Lewald, and H.-O. Karnath, ``Testing the dual-pathway
  model for auditory processing in human cortex,'' \emph{Neuroimage}, vol. 124,
  pp. 672--681, 2016.

\bibitem{AcousticFields}
A.~Luo, Y.~Du, M.~Tarr, J.~Tenenbaum, A.~Torralba, and C.~Gan, ``Learning
  neural acoustic fields,'' \emph{NIPS}, vol.~35, pp. 3165--3177, 2022.

\bibitem{cao2016interactive}
C.~Cao, Z.~Ren, C.~Schissler, D.~Manocha, and K.~Zhou, ``Interactive sound
  propagation with bidirectional path tracing,'' \emph{ACM Transactions on
  Graphics (TOG)}, vol.~35, no.~6, pp. 1--11, 2016.

\bibitem{veach1995bidirectional}
E.~Veach and L.~Guibas, ``Bidirectional estimators for light transport,'' in
  \emph{Photorealistic Rendering Techniques}.\hskip 1em plus 0.5em minus
  0.4em\relax Springer, 1995, pp. 145--167.

\bibitem{Matterport3D}
A.~Chang, A.~Dai, T.~Funkhouser, M.~Halber, M.~Niessner, M.~Savva, S.~Song,
  A.~Zeng, and Y.~Zhang, ``Matterport3d: Learning from rgb-d data in indoor
  environments,'' \emph{International Conference on 3D Vision (3DV)}, 2017.

\bibitem{chen2021semantic}
C.~Chen, Z.~Al-Halah, and K.~Grauman, ``Semantic audio-visual navigation,'' in
  \emph{CVPR}, 2021, pp. 15\,516--15\,525.

\bibitem{tatiya2021knowledge}
G.~Tatiya, J.~Francis, L.~Bondi, I.~Navarro, E.~Nyberg, J.~Sinapov, and J.~Oh,
  ``Knowledge-driven scene priors for semantic audio-visual embodied
  navigation,'' \emph{arXiv preprint arXiv:2212.11345}, 2022.

\bibitem{kartal2019terminal}
B.~Kartal, P.~Hernandez-Leal, and M.~E. Taylor, ``Terminal prediction as an
  auxiliary task for deep reinforcement learning,'' in \emph{AAAI}, vol.~15,
  no.~1, 2019, pp. 38--44.

\bibitem{hernandez2019agent}
P.~Hernandez-Leal, B.~Kartal, and M.~E. Taylor, ``Agent modeling as auxiliary
  task for deep reinforcement learning,'' in \emph{AAAI}, vol.~15, no.~1, 2019,
  pp. 31--37.

\bibitem{foerster2017stabilising}
J.~Foerster, N.~Nardelli, G.~Farquhar, T.~Afouras, P.~H. Torr, P.~Kohli, and
  S.~Whiteson, ``Stabilising experience replay for deep multi-agent
  reinforcement learning,'' in \emph{ICML}.\hskip 1em plus 0.5em minus
  0.4em\relax PMLR, 2017, pp. 1146--1155.

\bibitem{hong2018deep}
Z.-W. Hong, S.-Y. Su, T.-Y. Shann, Y.-H. Chang, and C.-Y. Lee, ``A deep policy
  inference q-network for multi-agent systems,'' in \emph{Proceedings of the
  17th International Conference on Autonomous Agents and MultiAgent Systems},
  2018, pp. 1388--1396.

\bibitem{liu2020return}
G.~Liu, C.~Zhang, L.~Zhao, T.~Qin, J.~Zhu, L.~Jian, N.~Yu, and T.-Y. Liu,
  ``Return-based contrastive representation learning for reinforcement
  learning,'' in \emph{ICLR}, 2020.

\bibitem{jaderberg2016reinforcement}
M.~Jaderberg, V.~Mnih, W.~M. Czarnecki, T.~Schaul, J.~Z. Leibo, D.~Silver, and
  K.~Kavukcuoglu, ``Reinforcement learning with unsupervised auxiliary tasks,''
  in \emph{ICLR}, 2017.

\bibitem{mirowski2019learning}
P.~Mirowski, ``Learning to navigate,'' in \emph{1st International Workshop on
  Multimodal Understanding and Learning for Embodied Applications}, 2019, pp.
  25--25.

\bibitem{2020VisualNavigation}
F.~Zeng, C.~Wang, and S.~S. Ge, ``A survey on visual navigation for artificial
  agents with deep reinforcement learning,'' \emph{IEEE Access}, vol.~8, pp.
  135\,426--135\,442, 2020.

\bibitem{makous1990two}
J.~C. Makous and J.~C. Middlebrooks, ``Two-dimensional sound localization by
  human listeners,'' \emph{The journal of the Acoustical Society of America},
  vol.~87, no.~5, pp. 2188--2200, 1990.

\bibitem{middlebrooks1991s}
J.~C. Middlebrooks and D.~M. Green, ``Sound localization by human listeners,''
  \emph{Annu. Rev. Psychol}, vol.~42, pp. 135--59, 1991.

\bibitem{PICINALI2014393}
L.~Picinali, A.~Afonso, M.~Denis, and B.~F. Katz, ``Exploration of
  architectural spaces by blind people using auditory virtual reality for the
  construction of spatial knowledge,'' \emph{International Journal of
  Human-Computer Studies}, vol.~72, no.~4, pp. 393--407, 2014.

\bibitem{morrongiello1990recent}
B.~A. Morrongiello and A.~Gotowiec, ``Recent advances in the behavioral study
  of infant audition: The development of sound localization skills.''
  \emph{Journal of Speech-Language Pathology and Audiology}, 1990.

\bibitem{morrongiello1994sound}
B.~A. Morrongiello, K.~D. Fenwick, L.~Hillier, and G.~Chance, ``Sound
  localization in newborn human infants,'' \emph{Developmental Psychobiology:
  The Journal of the International Society for Developmental Psychobiology},
  vol.~27, no.~8, pp. 519--538, 1994.

\bibitem{ganin2015unsupervised}
Y.~Ganin and V.~Lempitsky, ``Unsupervised domain adaptation by
  backpropagation,'' in \emph{ICML}.\hskip 1em plus 0.5em minus 0.4em\relax
  PMLR, 2015, pp. 1180--1189.

\bibitem{schulman2017proximal}
J.~Schulman, F.~Wolski, P.~Dhariwal, A.~Radford, and O.~Klimov, ``Proximal
  policy optimization algorithms,'' \emph{arXiv preprint arXiv:1707.06347},
  2017.

\bibitem{anderson2018evaluation}
P.~Anderson, A.~Chang, D.~S. Chaplot, A.~Dosovitskiy, S.~Gupta, V.~Koltun,
  J.~Kosecka, J.~Malik, R.~Mottaghi, M.~Savva \emph{et~al.}, ``On evaluation of
  embodied navigation agents,'' \emph{arXiv preprint arXiv:1807.06757}, 2018.

\bibitem{dijkstra_note_1959}
E.~W. Dijkstra, ``A note on two problems in connexion with graphs,''
  \emph{Numerische Mathematik}, vol.~1, no.~1, pp. 269--271, Dec. 1959.

\bibitem{majumder2022active}
S.~Majumder and K.~Grauman, ``Active audio-visual separation of dynamic sound
  sources,'' in \emph{ECCV}.\hskip 1em plus 0.5em minus 0.4em\relax Springer,
  2022, pp. 551--569.

\bibitem{tzinis2020two}
E.~Tzinis, S.~Venkataramani, Z.~Wang, C.~Subakan, and P.~Smaragdis, ``Two-step
  sound source separation: Training on learned latent targets,'' in
  \emph{ICASSP}.\hskip 1em plus 0.5em minus 0.4em\relax IEEE, 2020, pp. 31--35.

\bibitem{zhong2021towards}
F.~Zhong, P.~Sun, W.~Luo, T.~Yan, and Y.~Wang, ``Towards distraction-robust
  active visual tracking,'' in \emph{International Conference on Machine
  Learning}.\hskip 1em plus 0.5em minus 0.4em\relax PMLR, 2021, pp.
  12\,782--12\,792.

\end{thebibliography}
\clearpage
\newpage

\end{document}